\documentclass[runningheads]{llncs}

\usepackage[T1]{fontenc}
\usepackage[utf8]{inputenc}
\usepackage{xcolor}
\usepackage{tikz}
\usepackage{pgfplots}
\usepackage{listings}
\usetikzlibrary{arrows.meta, positioning, shapes,quotes}
\usetikzlibrary{arrows}
\usepackage{url}
\usepackage[hidelinks]{hyperref}
\usepackage{graphicx}
\usepackage{amsmath, amssymb}
\usepackage{booktabs} 
\usepackage{tabularx} 
\usepackage{array} 
\usepackage{ragged2e}
\usepackage{multirow}
\usepackage{placeins}
\ifdefined\DeclareUnicodeCharacter
\DeclareUnicodeCharacter{2606}{}
\DeclareUnicodeCharacter{2013}{--}
\fi
\pgfplotsset{compat=1.18}

\lstdefinestyle{compactprompt}{
    basicstyle=\ttfamily\fontsize{6.3pt}{6.6pt}\selectfont,
    breaklines=true,
    breakatwhitespace=false,
    columns=fullflexible,
    keepspaces=true,
    showstringspaces=false,
    aboveskip=0pt,
    belowskip=0pt,
    lineskip=-1pt,
    xleftmargin=0pt,
    xrightmargin=0pt,
    framexleftmargin=0pt,
    framexrightmargin=0pt
}
\lstdefinestyle{promptfigure}{
    basicstyle=\ttfamily\fontsize{7pt}{7.8pt}\selectfont,
    breaklines=true,
    breakatwhitespace=true,
    columns=fullflexible,
    keepspaces=true,
    showstringspaces=false,
    frame=single,
    rulecolor=\color{gray!45},
    backgroundcolor=\color{gray!6},
    xleftmargin=1mm,
    xrightmargin=1mm,
    framexleftmargin=1mm,
    framexrightmargin=1mm,
    aboveskip=2pt,
    belowskip=2pt
}

\newcommand{\embed}{{\sf E}}
\newcommand{\prompt}{{\sf Inst}}

\newcommand{\myabstract}{{\sf Tran}}

\newcommand{\BibTeX}{B\kern-.05em{\sc i\kern-.025em b}\kern-.08em\TeX}

\def\BibTeX{{\rm B\kern-.05em{\sc i\kern-.025em b}\kern-.08em
    T\kern-.1667em\lower.7ex\hbox{E}\kern-.125emX}}
\begin{document}

\title{Making Implicit Premises Explicit\\
in Logical Understanding of Enthymemes}
\titlerunning{Making Implicit Premises Explicit}

\author{Xuyao Feng\inst{1}\thanks{Corresponding author.} \and Anthony Hunter\inst{1}}
\authorrunning{X. Feng and A. Hunter}

\institute{Department of Computer Science, University College London, United Kingdom\\
\email{\{xuyao.feng.20,anthony.hunter\}@ucl.ac.uk}}
\maketitle

\begin{abstract}
Real-world arguments in text and dialogues are normally enthymemes (i.e. some of their premises and/or claims are implicit).
Natural language processing (NLP) methods for handling enthymemes can potentially identify enthymemes in text but they do not decode their underlying logic,
whereas logic-based approaches for handling them assume a knowledge base with sufficient formulae that can be used to decode them via abduction.
There is therefore a lack of a systematic method for translating textual components of an enthymeme into a logical argument and generating the logical formulae required for their decoding, and thereby showing logical entailment.
To address this, we propose a pipeline that integrates: (1) a large language model (LLM) to generate intermediate implicit premises based on the explicit premise and claim;
(2) a neural AMR parser based on a pretrained language model, followed by an AMR-to-logic translator, to convert the natural-language statements into logical formulas;
and (3) a neuro-symbolic reasoner based on a SAT solver to determine entailment. We evaluate the pipeline on two enthymeme datasets and report precision, recall, F1-score, and accuracy for selecting the correct implicit premise.

\end{abstract}
\section{Introduction}

Humans frequently engage in argumentation when confronted with conflicting evidence and opinions \cite{Atkinson2017}.
Computational models of argument provide formal languages for arguments and counterarguments
\cite{doi:10.1080/19462166.2013.869764}.
However, in the real-world, arguments are often incomplete (i.e., not all the premises or claims are explicit); these arguments are known as \textbf{enthymemes}. For example, for an argument with the claim {\em you should take an umbrella}, it may be insufficient to have the premise {\em the weather report predicts rain}, and so the logical connection between the premise and claim is implicit.

Existing work on enthymemes falls into two categories. The first category uses natural language processing (NLP) to identify and understand enthymemes in textual data \cite{Habernal.et.al.2018.NAACL.ARCT,doi:10.1177/19462174251344764}. The second category are symbolic approaches that use abduction to decode enthymemes: identifying missing premises which, when added to the explicit ones, logically entail the claim \cite{10.5555/1619645.1619657,Hunter_2022,ijcai2025p495}.
For example, given atoms $r$ ({\em the weather report predicts rain}) and $u$ ({\em take an umbrella}), the premise $\{r\}$ is insufficient to entail the claim $u$ (i.e. $\{r\} \nvdash u$). This enthymeme can be decoded by adding $r \rightarrow u$ to the premises to create a valid inference (i.e. $\{r, r \rightarrow u\} \vdash u$).

While the NLP approaches focus on the textual level and do not reconstruct the underlying logic-based argument structure, the symbolic approaches currently assume a sufficiently large set of logical formulas for decoding enthymemes but they do not consider how to obtain these formulas.
Therefore, there is a lack of a systematic method to first translate the textual components of a free-text enthymeme into a logical form and then use automated methods to decode it.

To address this shortcoming, we propose a neuro-symbolic pipeline that first utilizes an off-the-shelf large language model (LLM) to generate intermediate premises based on the given explicit premise and the claim. For example, for the premise {\em they add a lot to the piece and I look forward to reading comments}, and the claim {\em comment sections have not failed},
an intermediate premise could be {\em comments sections are a welcome distraction from my work}. So we identify intermediate premises that with the explicit premises can entail the claim.
The rest of the neuro-symbolic pipeline (summarized in Figure~\ref{fig:pipeline}) is based on translating the text (premises, implicit premises, and claim) into logic, and then using neuro-symbolic reasoning to show entailment (using word embeddings to identify predicates that can be treated as the same and thereby offers a form of commonsense reasoning that is more relaxed than reasoning with propositional logic) and so extends the proposal in \cite{Feng2025}.

We proceed as follows:
Section \ref{section:background} reviews necessary background;
Section \ref{section:pipeline} describes our pipeline;
Section \ref{section:evaluation} describes the datasets and evaluation;
Section \ref{section:results} discusses the results;
and Section \ref{section:discussion} discusses our contribution.

\section{Background}
\label{section:background}

This section reviews abstract meaning representation (AMR) and how we can translate it into propositional logic.

\subsection{Abstract meaning representation}
\label{section:amr}

Abstract meaning representation (AMR) is a semantic representation language for representing sentences as rooted, labelled, directed, and acyclic graphs (DAGs). AMR is intended to assign the same AMR graph to similar sentences, even if they are not identically worded. The approach was first introduced by Langkilde and Knight in 1998  \cite{langkilde-knight-1998-generation-exploits} as a derivation from the Penman Sentence Plan Language \cite{kasper-1989-flexible}. In 2013, AMRs re-gained attention due to Banarescu et al.  \cite{banarescu-etal-2013-abstract}, and were introduced into NLP tasks such as machine translation and natural language understanding. The modern (post-2010) AMR\footnote{\url{https://github.com/amrisi/amr-guidelines}} draws on predicate senses and semantic roles from the OntoNotes project  \cite{hovy-etal-2006-ontonotes}.

\begin{figure}
\footnotesize
    \[
\begin{array}{ll}
\mbox{\tt (w / want-01}
& \mbox{\tt (w / want-01}\\
\hspace{1cm}     \mbox{\tt    :arg0 (b / boy)}
& \hspace{1cm} \mbox{\tt :arg0 (b / boy)}\\
\hspace{1cm}   \mbox{\tt    :arg1 (g / go-01}
& \hspace{1cm} \mbox{\tt :arg1 (g / go-01}\\
\hspace{2cm}  \mbox{\tt          :arg0 b))}
& \hspace{2cm}\mbox{\tt :arg0 b}\\
&\hspace{2cm}\mbox{\tt  :polarity -))}
    \end{array}
    \]
    \caption{AMR for the sentence ``The boy wants to go.'' (left) and ``The boy does not want to go.'' (right).\\}
    \label{f1}
\end{figure}

In AMR, negation is represented via the :polarity relation. For example, Figure~\ref{f1} represents ``The boy does not want to go.''
In AMR, the numbers after the instance name (such as want-01 above) denote a particular OntoNotes or PropBank semantic frame  \cite{kingsbury-palmer-2002-treebank}. These frames have different parameters, but the subject is generally denoted by $\texttt{arg0}$ and the object by \texttt{arg1}.
The parameters, which draw out the semantic roles of the words in the AMR, include \texttt{location} (e.g. ``France''), \texttt{unit} (e.g. ``kilogrammes''), and \texttt{time} (e.g. ``yesterday'').
AMR uses the same structure to represent semantically similar texts by making several simplifying assumptions. AMR cannot represent verb tenses nor distinguish between verbs and nouns. Also it does not represent articles, quote marks, or the singular and plural.
Nonetheless, AMR offers a valuable formal abstraction of the meaning of sentences.



In our pipeline, we use the IBM Transition AMR parser\footnote{\url{https://github.com/IBM/transition-amr-parser/tree/master}} with the pretrained AMR 3.0 model AMR3-joint-ontowiki-seed43. The model was trained by Smatch-based ensemble distillation: Smatch measures graph overlap among candidate parses, and selected parses form silver data for training one parser \cite{DBLP:journals/corr/abs-2112-07790}.

We use AMR because its concepts and predicate--argument relations can be converted systematically into logic by the Bos algorithm. The pipeline therefore depends on AMR parsing accuracy and on the distinctions encoded by AMR.

\subsection{Translation of AMR into Logic}
\label{section:automatedreasoning}

The Bos algorithm \cite{bos-2016-squib} translates an AMR graph into a nested conjunction of atoms. A monadic (unary) atom has the form $p(a)$ and represents a concept or property. A dyadic (binary) atom has the form $r(a,b)$ and represents a relation between two concepts \cite{chanin2023neuro}.
The AMR for ``The boy does not want to go'' from Figure~\ref{f1} is converted into the following logical formula.
\[
\tt \exists w ( \exists b(want(w) \wedge arg0(w, b) \wedge boy(b) \wedge \neg \exists g (arg1(w, g) \wedge go(g) \wedge arg0(g, b))))
\]

There is an open-source Python library based on the Bos algorithm, the AMR-to-logic converter \cite{chanin2023neuro}, to translate AMR graphs into first-order logic. We extended this library, to give our AMR-to-propositional-logic translator, by rewriting each first-order logic formula to a propositional logic formula by grounding out the existentially quantified variables with new constants. We refer to each such propositional formula as an AMR formula.
For this, we make the assumption that for each existentially quantified variable, there is a specific entity that can represent the quantified variable. This can be viewed as a Skolem constant.
We choose each constant symbol for this grounding as follows: For each monadic predicate $r(a)$, we use $r$ as the constant symbol to replace the variable symbol $a$ throughout the formula.
This then means the monadic predicates are now redundant and so we delete them.
For the above example, the propositional logic formula is simplified as
${\tt arg0(want,boy)} \wedge \neg({\tt arg1(want,go)}\wedge {\tt arg0(go,boy)})$.

For our pipeline, we assume the usual definitions for propositional logic.
We start with a set of propositional atoms (letters), and we construct formulas in the usual way using the connectives for negation $\neg$, conjunction $\wedge$, disjunction $\lor$, implication $\leftarrow$, and biconditional $\leftrightarrow$.

\section{Pipeline}
\label{section:pipeline}

\begin{figure}[t]
\begin{center}
\begin{tikzpicture}[->,thick,
ARG/.style={draw,align=center, minimum width=88mm, minimum height=6mm,
shape=rectangle,
rounded corners=2pt,
fill=gray!20,font=\footnotesize}]
\node[font=\footnotesize] (a0)  at (0,7.8) {A set of sentences};
\node[ARG] (a00)  at (0,7) {Prompt an LLM for intermediate implicit premise(s)};
\node[ARG] (a1)  at (0,6) {Text-to-AMR Parser (based on a fine-tuned LLM)};
\node[ARG] (a2)  at (0,5) {AMR-to-Propositional-Logic Translator};
\node[ARG] (a3)  at (0,4) {Relaxation Methods (based on word embeddings and NLI)};
\node[ARG] (a4)  at (0,3) {Automated Reasoner (based on PySAT)};
\node[font=\footnotesize] (a5)  at (0,2.2) {Label (entailment and non-entailment)};
\path (a0)[] edge[] (a00);
\path (a00)[] edge[] (a1);
\path (a1)[] edge[] (a2);
\path (a2)[] edge[] (a3);
\path (a3)[] edge[] (a4);
\path (a4)[] edge[] (a5);
\end{tikzpicture}
\end{center}
\caption{Our neuro-symbolic pipeline where input is a set of natural language sentences (e.g. premise, implicit premises and claim), and the output is a label.}
\label{fig:pipeline}
\end{figure}

Our neuro-symbolic pipeline\footnote{\url{https://github.com/fxy-1117/SUM2026}},
which is summarized in Figure~\ref{fig:pipeline},
consists of five main components:
An LLM to generate intermediate implicit premises (Section \ref{llmg});
A neural text-to-AMR parser based on a pretrained language model (Section \ref{section:amr});
An AMR-to-propositional-logic translator (\ref{section:automatedreasoning});
A set of methods for relaxing the propositional formulas (\ref{section:similarity} and \ref{section:translate});
And an automated reasoner based on PySAT (\ref{section:classification}).


\subsection{Generate implicit premises}
\label{llmg}

\begin{figure}[t]
\begin{lstlisting}[style=promptfigure]
Generate two distinct chains of reasoning based on the premise and claim provided below. Follow these instructions carefully.
**Premise:** {premise}
**Claim:** {claim}
**Instructions:**
- **Helpful Chain:** Give one/two/three statements that represent the steps of reasoning starting from the premises and finishing with the claim (split by full stop), avoid using pronouns and repeating claim (each statement under 10 words)
- **Non-Helpful Chain:** Give one/two/three statements that represent the steps of reasoning starting from the premises and finishing with the neutral or contradictory of claim (split by full stop), avoid using pronouns and repeating claim (each statement under 10 words)
**Output Format:**
Your output must exactly match the following structure. Do not add any extra text, headers, or explanations.
Premise: {premise}
Claim: {claim}
Helpful: [insert helpful reasoning chain here]
Non-Helpful: [insert non-helpful reasoning chain here]
Replace the bracketed parts with your responses, ensuring each chain is a continuous sentence or phrase.
\end{lstlisting}
\caption{Prompt for generating helpful and unhelpful chains of
reasoning}
\label{prompt}
\end{figure}
We use an LLM to generate implicit premises by providing the premise and claim, then requesting one, two, or three steps of intermediate implicit premises.
The intermediate steps are intended to make explicit how the claim follows from the explicit premise (as a form of chain of reasoning \cite{Wei2022}). Furthermore, it is intended that as the number of steps is increased, more information is provided for how to show that the claim follows from the premises.
The LLM used is DeepSeek v3.2 \cite{deepseekai2025deepseekv32pushingfrontieropen}, and our prompt is presented in Figure~\ref{prompt}. The implicit premises, like the explicit premises and claims, are natural language statements, and so we use the text-to-AMR parser and the AMR-to-propositional-logic translator to obtain logical formulas which we handle in Sections \ref{section:similarity} and \ref{section:translate}.

For the evaluation (Section \ref{section:evaluation}), the implicit premises generated above are helpful premises. We used the same method to also generate unhelpful premises.  So for a premise and claim, the unhelpful premise together with the premise would contradict or be neutral with respect to the claim. So for the evaluation, we had one-, two-, and three-step helpful premises and one-, two-, and three-step unhelpful premises.

\subsection{Representing formulas}

An AMR formula is composed of a set of atoms together with the $\neg$ and $\wedge$ logical operators.
Since each atom is ground, an AMR formula is a propositional logic formula.

\begin{definition}
Let ${\cal A}$ be a set of AMR atoms (i.e. ground dyadic predicates of the form $r(a,b)$ where $a$ and $b$ are constant symbols).
The set of {\bf AMR formulas}, denoted ${\cal L}$, is defined inductively as follows:
If $\alpha\in{\cal A}$, then $\alpha\in{\cal L}$;
If $\alpha,\beta\in{\cal L}$, then $\alpha\wedge\beta\in{\cal L}$;
And if $\alpha\in{\cal L}$, then $\neg\alpha\in{\cal L}$.
\end{definition}

Given a natural language sentence $S$, our AMR-to-propositional-logic translator identifies an AMR formula $\phi$
that {\bf represents} $S$.

\begin{example}
\label{ex:amrformula}
The AMR formula $\tt \neg arg0(go,car)$
represents the sentence ``the car does not go''.
\end{example}

An abstract formula is also a propositional formula.

\begin{definition}
Let ${\cal P}$ be a set of propositional letters.
The set of {\bf abstract formulas}, denoted ${\cal F}$, is defined inductively as follows:
If $\alpha\in{\cal P}$, then $\alpha\in{\cal F}$;
If $\alpha,\beta\in{\cal F}$, then $\alpha\wedge\beta\in{\cal F}$;
And if $\alpha\in{\cal F}$, then $\neg\alpha\in{\cal F}$.
\end{definition}

If $|{\cal A}| = |{\cal P}|$, then for each $\alpha \in {\cal L}$, there is a $\beta\in {\cal F}$ (and vice versa) such that $\alpha$ and $\beta$ are isomorphic (i.e. they have the same syntax tree except for the atoms associated with the leaves). For example, $\tt \neg x_1$ is an abstract formula that is isomorphic to the AMR formula in Example \ref{ex:amrformula}.

Given an AMR formula, or an abstract formula, denoted $\phi$, let ${\sf Atoms}(\phi)$ denote the set of atoms used in $\phi$.

\subsection{Similarity measures}
\label{section:similarity}


In the next two subsections, we explain how we translate each AMR formula into an abstract formula,
which we then use with the PySAT automated reasoner.
For the translation, we check whether two atoms in an AMR formula can be regarded as equivalent, and therefore mapped to the same propositional letter in the corresponding abstract formula.
For this, if for two AMR atoms $\alpha$ and $\beta$,
the similarity between the two strings of words corresponding to the two atoms is greater
than a threshold $\tau_m$ (as explained below),
then we treat them as equivalent (denoted $\alpha\simeq\beta$) in the corresponding abstract formula,
and so it is a relaxation of the AMR formula.

An {\bf embedding} of a string $s$ is a vector $v=\embed(s)$ produced by an encoding function $\embed$.
We implement $\embed$ with the bge-small-en-v1.5 sentence-transformer model \cite{bge_embedding}.

The {\bf similarity} between two embedding vectors $v_1$ and $v_2$ is defined as follows, where $\theta$ is the angle between the vectors, $v_1 \cdot v_2$ is dot product between $v_1,v_2$, and $||v_1||$ represents the L2 norm.
\[
{\sf similarity}(v_1,v_2) = cos(\theta) = \frac{v_1 \cdot v_2}{||v_1||||v_2||}
\]

To obtain a string corresponding to an AMR atom, we use 25 relation-specific templates and one fallback template. Each relation-specific template maps an AMR predicate and its arguments to a string; the fallback concatenates the arguments. For example, the template ``[Y] [X]'' maps $\tt arg0(play,man)$ to ``{\em man play}''. The following definition formalizes this operation.


\begin{definition}
Let $\phi$ be an AMR formula,
and let $r(a,b)$ be an AMR atom in $\phi$ (i.e. $r(a,b) \in {\sf Atoms}(\phi)$).
Also let $T$ be a template for $r$ with placeholders $[{\rm X}]$ and $[{\rm Y}]$.
The {\bf instantiate function}, denoted $\prompt$, is defined as follows:
$\prompt(r(a,b), T)$ = $I$, where $I$ is the instantiation of $T$ in which $[{\rm X}]$ is replaced by $a$ and $[{\rm Y}]$ is replaced by $b$.
\end{definition}

Some examples of templates
are below where the AMR predicate name is given in brackets.

\begin{itemize}
\item ($\tt purpose$) ``[X] for [Y]''
\item ($\tt time$) ``[Y] is when action [X] takes place''
\item ($\tt arg1$) ``[X] [Y]''
\end{itemize}

The templates are fixed and shared across both datasets. Their wording can change the embedding similarities; we did not evaluate alternative wordings.

Next, we use the embedding function $\embed$ to obtain the sentence embeddings of the instantiations of templates.

\begin{example}
\label{e13}
Let $T$ = {\rm ``[Y] [X]''} be the template for AMR predicate name $\tt arg0$.
Therefore, for the AMR atom $\tt arg0(play,child)$,
$\prompt({\tt arg0(play,child)}, T)$ = ``child play''.
\end{example}



For the following definition of a matching relation, we assume that one AMR formula refers to a premise and the other refers to a claim.

\begin{definition}
Let AMR formula $\phi$ be a premise and AMR formula $\psi$ be a claim,
let $\alpha\in {\sf Atoms}(\psi)$ be an AMR atom of the form $r(a,b)$,
let $T_1$ be a template for $r$,
let $\beta\in {\sf Atoms}(\phi)$ be an AMR atom of the form $q(c,d)$,
let $T_2$ be a template for $q$,
let $\tau_m \in [0,1]$ be the {\bf neuro-matching threshold},
and let $\embed$ be an embedding function.
The {\bf neuro-matching relation}, denoted $\simeq$, is defined as follows
\[
\begin{array}{l}
\alpha \simeq \beta \mbox{ iff }  (\beta,x) \in {\sf Sim}(\alpha) \mbox{ and }\forall (\beta',y) \in {\sf Sim}(\alpha), x \geq y
\end{array}
\]
where
\[
\resizebox{\linewidth}{!}{$\displaystyle
{\sf Sim}(\alpha)=\{(\beta',x)\mid \beta'\in {\sf Atoms}(\phi),\ x > \tau_m,\ {\sf similarity}(\embed(\prompt(\alpha,T_1)),\embed(\prompt(\beta',T_2)))=x\}.
$}
\]
\end{definition}

For each claim atom $\alpha$, this definition selects the premise atom $\beta_i$ with the highest similarity above $\tau_m$; then $\alpha\simeq\beta_i$.
\begin{example}
\label{example:neuromatch}
For premise text ``A tiger is walking in the cage'', the premise atoms are ${\tt arg0(walk,tiger)}$ and ${\tt location(walk,cage)}$.
For claim text ``A tiger is moving'', the claim atom is ${\tt arg0(move,tiger)}$.
With template $T$ for $\tt arg0$ and $\tau_m=0.6$, the similarity score is $0.8483$, so ${\tt arg0(move,tiger)}\simeq{\tt arg0(walk,tiger)}$.
\end{example}
After identifying the best match for an AMR atom in a claim, we perform a contradiction check on this matched pair using a Natural Language Inference (NLI) model\footnote{mDeBERTa-v3-base-xnli-multilingual-nli-2mil7} \cite{Laurer2023}. The NLI model evaluates a pair of sentences \((S_1, S_2)\) by identifying three scores, each in the [0,100] interval:
\(s_{\text{Ent}}(S_1, S_2)\) is the degree to which \(S_2\) is entailed by \(S_1\);
\(s_{\text{Con}}(S_1, S_2)\) is the degree of conflict between \(S_1\) and \(S_2\);
and \(s_{\text{Neu}}(S_1, S_2)\) is the degree to which \(S_1\) and \(S_2\) are unrelated.
The model returns the label with the highest score (with a random choice in case of a tie).

\begin{definition}
Let \(\mathcal{S}\) be the set of all natural language sentences,
and let $\mathcal{C} = \{\text{Ent}, \text{Con}, \text{Neu}\}$
be the set of outcomes (entailment, contradiction, and neutral).
A {\bf natural language inference (NLI) function}
$\mathsf{N} : \mathcal{S} \times \mathcal{S} \rightarrow \mathcal{C}$
is defined as follows with scores $s_{\text{Ent}}$, $s_{\text{Con}}$, and $s_{\text{Neu}}$
for a given pair \((S_1, S_2)\).
\[
\mathsf{N}(S_1,S_2) = \arg\max_{\sigma \in \mathcal{C}} s_{\sigma}(S_1, S_2)
\]
\end{definition}



\begin{definition}
Let AMR formula $\phi$ be a premise and AMR formula $\psi$ be a claim,
let $\alpha\in {\sf Atoms}(\psi)$ be an AMR atom of the form $r(a,b)$,
let $T_1$ be a template for $r$,
let $\beta\in {\sf Atoms}(\phi)$ be an AMR atom of the form $q(c,d)$,
let $T_2$ be a template for $q$, let $\sf N$ be an NLI model,
and let $\tau_c \in [0,100]$ be the {\bf neuro-contradict threshold}. The {\bf neuro-contradict relation} $\perp$
is defined as follows
\[
\begin{array}{ll}
\alpha \perp \beta\ \mbox{iff}
& \mathsf{N}(\prompt(\alpha,T_1),\prompt(\beta,T_2)) = \text{Con} \mbox{ and } s_{\text{Con}}(\prompt(\alpha,T_1),\prompt(\beta,T_2)) \geq \tau_c
\end{array}
\]

\end{definition}
\begin{example}
\label{example:neuromatchc}
For premise text ``A tiger is walking in the cage'' and claim text ``The tiger is sleeping in the cage'', the premise atoms are ${\tt arg0(walk,tiger)}$ and ${\tt location(walk,cage)}$, and the claim atoms are ${\tt arg0(sleep,tiger)}$ and ${\tt location(sleep,cage)}$.
For templates $T_1$ and $T_2$ for $\tt arg0$ and $\tt location$, when $\tau_c=80$, the NLI labels are $\text{Con}$ with scores $85$ and $82$.
Hence ${\tt arg0(sleep,tiger)}\perp{\tt arg0(walk,tiger)}$ and ${\tt location(sleep,cage)}\perp{\tt location(walk,cage)}$.
\end{example}

The $\simeq$ relation is reflexive and symmetric, whereas the $\perp$ relation is irreflexive and symmetric.

\subsection{Translating AMR into abstract formulas}
\label{section:translate}

We now consider how we can translate each AMR formula into an abstract formula.
The first aim is to represent each AMR atom by an abstract atom of the form $\tt x_i$ in order to facilitate use by a PySAT solver, and the second aim is to take advantage of the neuro-matching and neuro-contradict relations to simplify and constrain the abstract formula.
For example, if we have a premise $\tt x_1 \wedge x_2 \wedge x_3$ and a claim $\tt x_1 \wedge x_4$ and $\tt x_3$ and $\tt x_4$ are very similar concepts, then we can change the claim to $\tt x_1 \wedge x_3$, and then show entailment holds using the CNF versions of these relaxed formulas with PySAT.
Similarly, if we have a premise $\tt x_5 \wedge x_6$ and a claim $\tt x_7$,
and the $\tt x_5 \bot x_7$ relationship holds,
then we can change the claim to $\tt \neg x_5$.

\begin{definition}
\label{def:translation}
Given a set of AMR formulas $\Phi$,
and a set of neuro-matching and neuro-contradict relationships $\Psi$,
the function $g: {\cal A} \rightarrow {\cal P} \cup \{ \neg x \mid x \in {\cal P} \}$ is a {\bf mapping} for $\Phi$ and $\Psi$
iff for all $\phi,\phi'\in\Phi$,
for all $\alpha\in {\sf Atoms}(\phi)$,
for all $\beta\in {\sf Atoms}(\phi')$,
(1) $\alpha \simeq \beta\in\Psi  \mbox{ iff } g(\alpha) = g(\beta)$;
and (2) $\alpha \perp \beta\in\Psi  \mbox{ iff } g(\alpha) = \neg g(\beta)$.
For tautology $\top$, $g(\top) = \top$.
\end{definition}

The above definition ensures that if there are similar atoms, according to $\simeq$,
(resp. contradictory atoms, according to $\bot$),
then they are translated to the same atom (resp. complementary literals) in the abstract formulas.

\begin{example}
\label{ex:translation}
Let $a={\tt arg1(car,red)}$ and $b={\tt arg2(car,fast)}$.
Let $\phi_1=a\wedge b$, $\phi_2=a$, and $\Phi=\{\phi_1,\phi_2\}$.
A mapping is $g(a)={\tt x_1}$ and $g(b)={\tt x_2}$.
\end{example}

\begin{example}
\label{e7}
In Ex.~\ref{example:neuromatch}, $a={\tt arg0(move,tiger)}$,
$b={\tt arg0(walk,tiger)}$, and $c={\tt location(walk,cage)}$.
Then $g(a)=g(b)={\tt x_1}$ and $g(c)={\tt x_2}$.
\end{example}

\begin{example}
\label{e8}
In Ex.~\ref{example:neuromatchc}, $a={\tt arg0(walk,tiger)}$,
$b={\tt arg0(sleep,tiger)}$, $c={\tt location(walk,cage)}$, and
$d={\tt location(sleep,cage)}$.
Then $g(a)={\tt x_1}$, $g(b)=\neg{\tt x_1}$, $g(c)={\tt x_2}$, and $g(d)=\neg{\tt x_2}$.
\end{example}

Next we specify how an AMR formula is translated into an abstract formula using a mapping function.

\begin{definition}
Let $g$ be a mapping for a set of AMR formulas $\Phi$ and a set of neuro-matching and neuro-contradict relationships $\Psi$.
For $\phi \in \Phi$, a {\bf translation} of $\phi$ is
$\myabstract_g(\phi)$ where $\myabstract_g$ is defined as:
(1) $\myabstract_g(\alpha\wedge\beta) = \myabstract_g(\alpha) \wedge \myabstract_g(\beta)$;
(2) $\myabstract_g(\neg\alpha) = \neg\myabstract_g(\alpha)$;
And (3) $\myabstract_g(\alpha) = g(\alpha)$ when $\alpha\in{\cal A}$.
For $\Phi$, let $\myabstract_g(\Phi)$ = $\{\myabstract_g(\phi) \mid \phi \in \Phi\}$.
\end{definition}

\begin{example}
Continuing Example \ref{ex:translation},
$\myabstract_g(\phi_1)={\tt x_1}\wedge{\tt x_2}$ and $\myabstract_g(\phi_2)={\tt x_1}$.
\end{example}

So use of embeddings and NLI allow for identification of neuro-matching and neuro-contradict relations, which can then be used to rewrite AMR formulas into abstract formulas, where the latter are relaxations of the former.
So our relaxation methods rewrite the AMR formulas (output from the AMR-to-propositional-logic translator) into abstract formulas for use in automated reasoning as described next.


\subsection{Automated Reasoning}
\label{section:classification}

For the automated reasoning, we transform all the abstract formulas (from the previous section) into conjunctive normal form (CNF) using SymPY \cite{10.7717/peerj-cs.103}.
Then, we use PySAT \cite{imms-sat18}, which integrates several widely used state-of-the-art SAT solvers as theorem provers to check whether a CNF is consistent.


The aim of our pipeline is to identify the relationship between a premise $\phi$ (which may be a conjunction of explicit premise and one or more intermediate premises) and a claim $\psi$.  To prove whether $\phi$ entails $\psi$, we need to determine whether $\{\phi\} \vdash \psi$ holds. This is equal to determining whether $\phi \land\neg \psi$ is inconsistent. To do this, we need to change $\phi \land\neg \psi$ into a CNF formula, and then we can directly use PySAT to check consistency. Similarly, to prove whether $\phi$ contradicts $\psi$, we need to determine whether $\{\phi, \psi\} \vdash \perp$ holds. To do this, we need to change
$\phi\land\psi$ into a CNF formula, and then we can directly use PySAT to check consistency.

\section{Datasets and Evaluation}
\label{section:evaluation}

To evaluate our pipeline, we used two datasets (ARCT and ANLI) which we describe next.

\paragraph{ARCT} The Argument Reasoning Comprehension Task dataset (ARCT) is a benchmark dataset designed to evaluate the ability of an NLP model to identify the implicit reasoning (warrant) connecting a claim to its supporting premise \cite{Habernal.et.al.2018.NAACL.ARCT}.
Each data instance consists of a claim, a premise, two candidate implicit premises, where one is helpful (correct) and one is unhelpful (incorrect), and a binary label indicating which is the helpful premise. The task requires selecting which premise correctly justifies why the premise supports the claim.


\paragraph{ANLI} The Abductive Natural Language Inference dataset (ANLI), an Abductive Commonsense Reasoning challenge, evaluates the ability of an NLP model to use abductive reasoning for inferring the most plausible explanation given incomplete observations \cite{DBLP:journals/corr/abs-1908-05739}. Each instance presents an incomplete story beginning with two observations (O1 and O2 s.t. O2 happens after O1)  and provides two possible hypotheses (H1 and H2) that bridge the narrative gap. The task is to select the hypothesis that gives the most coherent/plausible explanation for the events. We treat O1 (resp. O2) as the explicit premise (resp. claim), and treat the most coherent/plausible explanation (resp. the other explanation) as the helpful (resp. unhelpful) implicit premise.

\begin{table}[t]
\centering
\caption{Examples from the ARCT and ANLI datasets}
\label{tab:example:ARCT}
\label{tab:example:ANLI}
\footnotesize
\begin{tabularx}{\linewidth}{@{}>{\bfseries}p{13mm}>{\RaggedRight\arraybackslash}X@{\hspace{1em}}>{\bfseries}p{13mm}>{\RaggedRight\arraybackslash}X@{}}
\toprule
\multicolumn{2}{c}{\textbf{ARCT}} & \multicolumn{2}{c}{\textbf{ANLI}} \\
\cmidrule(r{1em}){1-2}\cmidrule(l){3-4}
Premise & They add a lot to the piece and I look forward to reading comments.
& Premise & Jane was a professor teaching piano to students. \\
Claim & Comment sections have not failed.
& Claim & Jane spent the morning sipping coffee and reading a book. \\
Helpful & Comments sections are a welcome distraction from my work.
& Helpful & None of Jane's students had a lesson that day. \\
Unhelp. & Comments sections always distract me from my work.
& Unhelp. & Two of Jane's students were early for their lessons. \\
\bottomrule
\end{tabularx}
\end{table}

We augmented both datasets with one-, two-, and three-step helpful and unhelpful chains using the method in Section~\ref{llmg}. During evaluation, generation was disabled and the chains were read from the fixed augmented data. Each source item produced two binary examples: the premise, helpful chain, and claim with an entailment label; and the premise, unhelpful chain, and claim with a non-entailment label.

We used this transformed data in our neuro-symbolic pipeline, where $\phi$ was the logical formula of the conjunction of the premise and implicit premise(s), and $\psi$ was the logical formula of the claim. Thus, \textbf{entailment} (resp. \textbf{non-entailment}) meant the prediction was entailment  (resp. either contradiction or neutral).

Both benchmarks provide two candidate implicit premises. The evaluation therefore tests classification of fixed candidates, not unrestricted premise generation.

\section{Results}
\label{section:results}

\begin{table}[t]
  \centering
  \caption{Best F1-scores and thresholds by dataset, class (0 non-entailment, 1 entailment), and step type.}
  \label{tab:best_f1_scores}
  \footnotesize
  \setlength{\tabcolsep}{3pt}
  \begin{tabular}{@{}llccc@{\hspace{1.2em}}llccc@{}}
    \toprule
    \multicolumn{5}{c}{ANLI} & \multicolumn{5}{c}{ARCT} \\
    \cmidrule(r){1-5}\cmidrule(l){6-10}
    Class & Step type & F1 & $\tau_m$ & $\tau_c$ &
    Class & Step type & F1 & $\tau_m$ & $\tau_c$ \\
    \midrule
    \multirow{4}{*}{0} & original & 0.67 & 0.8 & 80
    & \multirow{4}{*}{0} & original & 0.67 & 0.5 & 90 \\
    & 1-step & 0.72 & 0.6 & 80
    & & 1-step & 0.72 & 0.7 & 90 \\
    & 2-step & 0.74 & 0.5 & 80
    & & 2-step & 0.70 & 0.75 & 80 \\
    & 3-step & 0.75 & 0.6 & 80
    & & 3-step & 0.72 & 0.7 & 90 \\
    \cmidrule(r){1-5}\cmidrule(l){6-10}
    \multirow{4}{*}{1} & original & 0.46 & 0.55 & 100
    & \multirow{4}{*}{1} & original & 0.21 & 0.55 & 80 \\
    & 1-step & 0.57 & 0.55 & 100
    & & 1-step & 0.45 & 0.65 & 90 \\
    & 2-step & 0.66 & 0.6 & 100
    & & 2-step & 0.52 & 0.6 & 90 \\
    & 3-step & 0.67 & 0.55 & 90
    & & 3-step & 0.58 & 0.6 & 90 \\
    \bottomrule
  \end{tabular}
\end{table}
\begin{table}[t]
\centering
\caption{Entailment accuracy by dataset and step type. None means no implicit premise; original means the dataset premise; 1-, 2-, and 3-step mean LLM-generated intermediate premises.}
\label{tab:ent}
\begin{tabular}{@{}lccccc@{}}
\toprule
\textbf{Dataset} & \textbf{none} & \textbf{original} & \textbf{1-step} & \textbf{2-step} & \textbf{3-step} \\
\midrule
\textbf{ANLI} & 0.496 & 0.520 & 0.624 & 0.660 & 0.716 \\
\textbf{ARCT} & 0.276 & 0.292 & 0.480 & 0.496 & 0.560 \\
\bottomrule
\end{tabular}
\end{table}

For each dataset, we evaluated neuro-matching thresholds \(\tau_m\) from 0.5 to 1 and neuro-contradict thresholds \(\tau_c\) of 80, 90, and 100. For each dataset, step type, and threshold pair, we shuffled the fixed data pool with seed 1129 and processed rows until 150 valid source rows were counted. Each counted row produced one example per class. We skipped rows when the premise alone entailed the claim, either candidate returned both entailment and contradiction, or evaluation raised an exception. When the pool was exhausted, the support was lower. At the best-F1 points for ARCT original, the class-0 and class-1 supports were 121 and 120, respectively; all other best-F1 points reached 150 per class.

\begin{figure}[!t]
  \centering
\pgfplotsset{
  accuracyplot/.style={
    width=0.8\linewidth,
    height=48mm,
    scale only axis,
    xlabel={Neuro-Matching Threshold ($\tau_m$)},
    ylabel={Accuracy},
    tick label style={font=\footnotesize},
    label style={font=\footnotesize},
    title style={font=\small},
    axis line style={line width=0.45pt},
    tick style={line width=0.45pt},
    every axis plot/.append style={line width=0.9pt, mark size=1.7pt},
    grid=major,
    ymin=0.45,
    ymax=0.75,
    xtick={0.5,0.6,0.7,0.8,0.9,1.0},
    ymajorgrids=true,
    major grid style={line width=0.15pt,draw=gray!35},
  }
}
  \begin{minipage}[t]{0.495\linewidth}
      \centering
      \begin{tikzpicture}
\begin{axis}[
    accuracyplot,
    title={ANLI},
]

\addplot[color=blue, mark=*, thick] coordinates {
    (0.5,0.510000) (0.55,0.516667) (0.6,0.486667) (0.65,0.493333) (0.7,0.503333) (0.75,0.500000) (0.8,0.500000) (0.85,0.500000) (0.9,0.500000) (0.95,0.500000) (1,0.500000)
};

\addplot[color=blue, mark=square*, thick] coordinates {
    (0.5,0.513333) (0.55,0.503333) (0.6,0.486667) (0.65,0.493333) (0.7,0.506667) (0.75,0.500000) (0.8,0.500000) (0.85,0.500000) (0.9,0.500000) (0.95,0.500000) (1,0.500000)
};

\addplot[color=blue, mark=triangle*, thick] coordinates {
    (0.5,0.513333) (0.55,0.506667) (0.6,0.500000) (0.65,0.503333) (0.7,0.506667) (0.75,0.496667) (0.8,0.500000) (0.85,0.500000) (0.9,0.500000) (0.95,0.500000) (1,0.500000)
};

\addplot[color=orange, mark=*, thick] coordinates {
    (0.5,0.620000) (0.55,0.636667) (0.6,0.643333) (0.65,0.603333) (0.7,0.550000) (0.75,0.523333) (0.8,0.513333) (0.85,0.506667) (0.9,0.500000) (0.95,0.500000) (1,0.500000)
};

\addplot[color=orange, mark=square*, thick] coordinates {
    (0.5,0.623333) (0.55,0.633333) (0.6,0.653333) (0.65,0.610000) (0.7,0.553333) (0.75,0.523333) (0.8,0.513333) (0.85,0.506667) (0.9,0.500000) (0.95,0.500000) (1,0.500000)
};

\addplot[color=orange, mark=triangle*, thick] coordinates {
    (0.5,0.570000) (0.55,0.596667) (0.6,0.616667) (0.65,0.603333) (0.7,0.550000) (0.75,0.520000) (0.8,0.513333) (0.85,0.506667) (0.9,0.500000) (0.95,0.500000) (1,0.500000)
};

\addplot[color=red, mark=*, thick] coordinates {
    (0.5,0.680000) (0.55,0.683333) (0.6,0.676667) (0.65,0.613333) (0.7,0.563333) (0.75,0.530000) (0.8,0.513333) (0.85,0.510000) (0.9,0.506667) (0.95,0.503333) (1,0.503333)
};

\addplot[color=red, mark=square*, thick] coordinates {
    (0.5,0.670000) (0.55,0.676667) (0.6,0.670000) (0.65,0.613333) (0.7,0.556667) (0.75,0.526667) (0.8,0.513333) (0.85,0.510000) (0.9,0.506667) (0.95,0.503333) (1,0.503333)
};

\addplot[color=red, mark=triangle*, thick] coordinates {
    (0.5,0.570000) (0.55,0.626667) (0.6,0.680000) (0.65,0.633333) (0.7,0.580000) (0.75,0.533333) (0.8,0.520000) (0.85,0.510000) (0.9,0.506667) (0.95,0.503333) (1,0.503333)
};

\addplot[color=green, mark=*, thick] coordinates {
    (0.5,0.703333) (0.55,0.713333) (0.6,0.703333) (0.65,0.666667) (0.7,0.600000) (0.75,0.536667) (0.8,0.520000) (0.85,0.506667) (0.9,0.510000) (0.95,0.510000) (1,0.506667)
};

\addplot[color=green, mark=square*, thick] coordinates {
    (0.5,0.696667) (0.55,0.713333) (0.6,0.700000) (0.65,0.660000) (0.7,0.603333) (0.75,0.540000) (0.8,0.520000) (0.85,0.506667) (0.9,0.510000) (0.95,0.510000) (1,0.506667)
};

\addplot[color=green, mark=triangle*, thick] coordinates {
    (0.5,0.566667) (0.55,0.643333) (0.6,0.653333) (0.65,0.676667) (0.7,0.596667) (0.75,0.546667) (0.8,0.516667) (0.85,0.506667) (0.9,0.510000) (0.95,0.510000) (1,0.506667)
};

\end{axis}
\end{tikzpicture}
  \end{minipage}
  \hfill
  \begin{minipage}[t]{0.495\linewidth}
      \centering
      \begin{tikzpicture}
\begin{axis}[
    accuracyplot,
    title={ARCT},
]

\addplot[color=blue, mark=*, thick] coordinates {
    (0.5,0.527273) (0.55,0.516667) (0.6,0.510204) (0.65,0.506667) (0.7,0.513333) (0.75,0.513333) (0.8,0.500000) (0.85,0.500000) (0.9,0.500000) (0.95,0.500000) (1,0.500000)
};

\addplot[color=blue, mark=square*, thick] coordinates {
    (0.5,0.533058) (0.55,0.522901) (0.6,0.516667) (0.65,0.503333) (0.7,0.516667) (0.75,0.513333) (0.8,0.500000) (0.85,0.500000) (0.9,0.500000) (0.95,0.500000) (1,0.500000)
};

\addplot[color=blue, mark=triangle*, thick] coordinates {
    (0.5,0.496667) (0.55,0.496667) (0.6,0.506667) (0.65,0.503333) (0.7,0.503333) (0.75,0.496667) (0.8,0.500000) (0.85,0.500000) (0.9,0.500000) (0.95,0.500000) (1,0.500000)
};

\addplot[color=orange, mark=*, thick] coordinates {
    (0.5,0.580000) (0.55,0.583333) (0.6,0.596667) (0.65,0.606667) (0.7,0.606667) (0.75,0.573333) (0.8,0.546667) (0.85,0.526667) (0.9,0.506667) (0.95,0.503333) (1,0.503333)
};

\addplot[color=orange, mark=square*, thick] coordinates {
    (0.5,0.593333) (0.55,0.593333) (0.6,0.606667) (0.65,0.623333) (0.7,0.620000) (0.75,0.580000) (0.8,0.546667) (0.85,0.530000) (0.9,0.506667) (0.95,0.503333) (1,0.503333)
};

\addplot[color=orange, mark=triangle*, thick] coordinates {
    (0.5,0.556667) (0.55,0.556667) (0.6,0.553333) (0.65,0.566667) (0.7,0.573333) (0.75,0.540000) (0.8,0.516667) (0.85,0.516667) (0.9,0.506667) (0.95,0.503333) (1,0.503333)
};

\addplot[color=red, mark=*, thick] coordinates {
    (0.5,0.584821) (0.55,0.592437) (0.6,0.602740) (0.65,0.613333) (0.7,0.606667) (0.75,0.600000) (0.8,0.573333) (0.85,0.536667) (0.9,0.503333) (0.95,0.503333) (1,0.503333)
};

\addplot[color=red, mark=square*, thick] coordinates {
    (0.5,0.599174) (0.55,0.604651) (0.6,0.620000) (0.65,0.623333) (0.7,0.613333) (0.75,0.600000) (0.8,0.573333) (0.85,0.536667) (0.9,0.503333) (0.95,0.503333) (1,0.503333)
};

\addplot[color=red, mark=triangle*, thick] coordinates {
    (0.5,0.540000) (0.55,0.533333) (0.6,0.570000) (0.65,0.580000) (0.7,0.576667) (0.75,0.563333) (0.8,0.546667) (0.85,0.520000) (0.9,0.503333) (0.95,0.503333) (1,0.503333)
};

\addplot[color=green, mark=*, thick] coordinates {
    (0.5,0.621622) (0.55,0.634454) (0.6,0.650000) (0.65,0.646667) (0.7,0.656667) (0.75,0.636667) (0.8,0.616667) (0.85,0.563333) (0.9,0.520000) (0.95,0.513333) (1,0.513333)
};

\addplot[color=green, mark=square*, thick] coordinates {
    (0.5,0.632231) (0.55,0.643411) (0.6,0.660000) (0.65,0.656667) (0.7,0.653333) (0.75,0.633333) (0.8,0.613333) (0.85,0.563333) (0.9,0.520000) (0.95,0.513333) (1,0.513333)
};

\addplot[color=green, mark=triangle*, thick] coordinates {
    (0.5,0.543333) (0.55,0.556667) (0.6,0.566667) (0.65,0.583333) (0.7,0.610000) (0.75,0.593333) (0.8,0.563333) (0.85,0.540000) (0.9,0.520000) (0.95,0.513333) (1,0.513333)
};

\end{axis}
\end{tikzpicture}
  \end{minipage}
  \par\vspace{0.8mm}
  {\scriptsize
  \begin{tabular}{@{}llll@{}}
  \textcolor{blue}{\rule{8mm}{0.9pt}} original &
  \textcolor{orange}{\rule{8mm}{0.9pt}} 1-step &
  \textcolor{red}{\rule{8mm}{0.9pt}} 2-step &
  \textcolor{green}{\rule{8mm}{0.9pt}} 3-step\\[-0.3mm]
  \(\bullet\) $\tau_c=80$ &
  \(\blacksquare\) $\tau_c=90$ &
  \(\blacktriangle\) $\tau_c=100$ &
  \end{tabular}}
  \caption{Overall accuracy for ANLI and ARCT by step type and thresholds. Original uses dataset premises; 1-, 2-, and 3-step use LLM-generated premises.}
  \label{anlifig}
\end{figure}

Figure~\ref{anlifig} shows the parameter sweep. Three-step premises reach the highest peak accuracy on both datasets, with the peaks between $\tau_m=0.55$ and $0.7$. Values of $\tau_c=80$ or $90$ produce higher peak accuracy than $\tau_c=100$. Table~\ref{tab:best_f1_scores} reports the best F1-score for each class and step type.

For entailment accuracy, Table~\ref{tab:ent} compares the no-premise ablation, the original helpful premise, and generated 1-, 2-, and 3-step premises on 250 entailment items per setting. Thresholds are \((\tau_m,\tau_c)=(0.55,90)\) for ANLI and \((0.65,90)\) for ARCT. Accuracy increases with the number of generated steps, and generated 1-step premises outperform the original premises.

\begin{figure}[t]
\begin{center}
\begin{tikzpicture}[scale=0.58, transform shape,
    node/.style={rectangle, draw, rounded corners, align=center, line width=1.2pt},
    claim/.style={
        node,
        fill=gray!10,
        rectangle split,
        rectangle split parts=2,
        font=\scriptsize,
        every split part/.style={dashed},
        text width=4cm
    },
    premise/.style={
        node,
        fill=gray!10,
        rectangle split,
        rectangle split parts=2,
        font=\scriptsize,
        every split part/.style={draw, dashed},
        text width=4cm
    },
    condition/.style={node, fill=white, text width=5.8cm, font=\scriptsize, minimum height=1cm},
    condition1/.style={node, fill=white, text width=1.2cm, font=\scriptsize, minimum height=1cm},
    condition2/.style={node, fill=white, text width=6.7cm, font=\scriptsize, minimum height=1cm},
    attack/.style={->, black, line width=1.2pt, >=stealth},
    attack1/.style={->, red, line width=1.2pt, >=stealth},
    support/.style={->, blue, line width=1.2pt, >=stealth},
    neutral/.style={->, green, line width=1.2pt, >=stealth}
]
\node[claim, text width=50mm] (main1) at (7.5,2)
{   \textbf{Implicit Premise A}: Torn webs result from trapped prey escaping.
    \nodepart{second}
    $ \tt arg1(result, escape) \wedge arg0(escape, prey) \wedge arg1(trap, prey) \wedge arg2(result, web) \wedge arg1(tear, web)$};
\node[claim, text width=50mm] (main2) at (-7,0)
{   \textbf{Implicit Premise B}: Small insect fled.
    \nodepart{second}
    $ \tt arg0(flee, small\ insect) $};
\node[claim, text width=50mm] (main3) at (-2,2.5)
{   \textbf{Implicit Premise C}: Wind tears a spiderweb.
    \nodepart{second}
    $ \tt arg1(tear, spiderweb) \wedge arg0(wind, tear)$};
\node[claim, text width=60mm] (counter) at (0,7)
{   \textbf{Claim}: A large insect escaped recently.
    \nodepart{second}
    $\tt arg0(escape, large\ insect) \wedge time(escape, recent)$};
\node[premise, text width=35mm] (p1) at (0,0)
{ \textbf{Premise}: The spiderweb is torn.
    \nodepart{second}
    $\tt arg1(tear, spiderweb)$};

\draw[support] (main1) edge[edge node={node[condition, pos=1.5] {\textbf{Support}\\ $\tt arg0(escape, large\ insect) \simeq \tt arg0(escape, prey)$ \\ $\tt time(escape, recent) \simeq \tt arg0(escape, prey)$}}] (7.5,4) to (7.5,7) to (counter);
\draw[attack1] (main2) edge[edge node={node[condition2, pos=1.1] {\textbf{Contradict}\\ $\tt arg0(escape, large\ insect) \perp \tt arg0(flee, small\ insect)$}},] (-7,4) to (-7,7) to (counter);
\draw[neutral] (p1) to (3,1) to [edge node={node[condition1, pos=0.7] {\textbf{Neutral}}}] (3,4) to (3,5) to (counter);
\draw[neutral] (main3) edge[edge node={node[condition1, pos=0.5] {\textbf{Neutral}}}] (counter);
\draw[attack] (p1) to (6,0) to (main1);
\draw[attack] (p1) to (main2);
\draw[attack] (p1) to (main3);
\end{tikzpicture}
\end{center}
\caption{A structured argument graph for decoding an enthymeme. The claim that a large insect escaped recently is supported by Implicit Premise A (torn webs result from trapped prey escaping), contradicted by Implicit Premise B (a small insect fled), and neutral with respect to Implicit Premise C (wind tears a spiderweb) and the premise alone. Arc labels show the neuro-matching relations used for support and the neuro-contradict relation used for contradiction. Thus, adding the AMR formulae for Implicit Premise A to those for the premise, together with the neuro-matching relations, entails the claim. Black arrows denote combining the explicit premise with an implicit premise.}
\label{fig:argument-graph}
\end{figure}

\FloatBarrier
\section{Discussion}
\label{section:discussion}

We presented a neuro-symbolic pipeline that generates intermediate premises, converts them to logic, and tests entailment with relaxed logical reasoning. On ARCT and ANLI, generated multi-step premises increase entailment accuracy, and entailment F1 remains lower than non-entailment F1 on ARCT. The pipeline exposes the generated formulas and relaxation relations, as illustrated in Figure~\ref{fig:argument-graph}.

The experiments compare two fixed candidate premises and do not evaluate unrestricted generation or external classifiers. They measure the effect of premise depth within the pipeline, not state-of-the-art classification performance. Applying the pipeline to another domain requires explicit premises, candidate implicit premises, and claims, and depends on AMR parsing and template coverage. Future work will evaluate other semantic representations, template sets, and unrestricted premise generation.

\subsubsection*{Disclosure of Interests.}
The authors have no competing interests to declare that are relevant to the content of this article.

\bibliographystyle{splncs04}
\bibliography{IEEEabrv,references}

\end{document}